\documentclass[letterpaper]{article} 
\usepackage{aaai23}  
\usepackage{times}  
\usepackage{helvet}  
\usepackage{courier}  
\usepackage[hyphens]{url}  
\usepackage{graphicx} 
\usepackage{natbib}  
\usepackage{caption} 
\usepackage{algorithm}
\usepackage{algorithmic}
\usepackage{tikz}

\def\checkmark{\tikz\fill[scale=0.25](0,.35) -- (.25,0) -- (1,.7) -- (.25,.15) -- cycle;} 

\usepackage{enumitem}
\usepackage{xcolor}
\usepackage{amsmath}
\usepackage{threeparttable}
\usepackage{multirow}
\usepackage{placeins}

\newcommand{\eqqref}[1]{Eq.~\ref{#1}}
\newcommand{\figref}[1]{Fig.~\ref{#1}}
\newcommand{\tableref}[1]{Table~\ref{#1}}
\newcommand{\algoref}[1]{Alg~\ref{#1}}

\usepackage{newfloat}
\usepackage{listings}
\DeclareCaptionStyle{ruled}{labelfont=normalfont,labelsep=colon,strut=off} 
\floatstyle{ruled}
\newfloat{listing}{tb}{lst}{}
\floatname{listing}{Listing}
\title{Bidirectional Domain Mixup for Domain Adaptive Semantic Segmentation}
\author{
    Daehan Kim\textsuperscript\equalcontrib\textsuperscript{\rm 1},
    Minseok Seo\textsuperscript\equalcontrib\textsuperscript{\rm 2},
    Kwanyong Park\textsuperscript{\rm 3},
    Inkyu Shin\textsuperscript{\rm 3},
    Sanghyun Woo\textsuperscript{\rm 3},
    In-So Kweon\textsuperscript{\rm 3},
    Dong-Geol Choi\corresp\textsuperscript{\rm 1}
}
\affiliations{
    \textsuperscript{\rm 1} Hanbat National University, Korea\\
    \textsuperscript{\rm 2} SI Analytics, Korea\\
    \textsuperscript{\rm 3} Korea Advanced Institute of Science and Technology (KAIST), Korea\\

    daehan.kim@edu.hanbat.ac.kr,
    minseok.seo@si-analytics.ai,
    \{pkyong7, dlsrbgg33, shwoo93, iskweon77\}@kaist.ac.kr
    dgchoi@hanbat.ac.kr
}

\usepackage{bibentry}

\begin{document}

\maketitle

\begin{abstract}
Mixup provides interpolated training samples and allows the model to obtain smoother decision boundaries for better generalization.
The idea can be naturally applied to the domain adaptation task, where we can mix the source and target samples to obtain domain-mixed samples for better adaptation. 
However, the extension of the idea from classification to segmentation (i.e., structured output) is nontrivial.
This paper systematically studies the impact of mixup under the domain adaptaive semantic segmentation task and presents a simple yet effective mixup strategy called Bidirectional Domain Mixup (BDM).
In specific, we achieve domain mixup in two-step: cut and paste.
Given the warm-up model trained from any adaptation techniques, we forward the source and target samples and perform a simple threshold-based cut out of the unconfident regions (\textbf{cut}).
After then, we fill-in the dropped regions with the other domain region patches (\textbf{paste}).
In doing so, we jointly consider class distribution, spatial structure, and pseudo label confidence.
Based on our analysis, we found that BDM leaves domain transferable regions by \textbf{cutting},
balances the dataset-level class distribution while preserving natural scene context by \textbf{pasting}.
We coupled our proposal with various state-of-the-art adaptation models and observe significant improvement consistently.
We also provide extensive ablation experiments to empirically verify our main components of the framework.
Visit our project page with the code at \url{https://sites.google.com/view/bidirectional-domain-mixup}.
\end{abstract}

\section{Introduction}

To reduce the annotation budget in semantic segmentation that require pixel level annotation, there have been many domain adaptation (DA) approaches using relatively inexpensive source (\textit{e.g.} simulator-based) data~\cite{richter2016playing, ros2016synthia} and unlabeled target (\textit{e.g.} real) data. 
However, deep neural networks show poor generalization performance in real data because they are sensitive to domain misalignments such as layout~\cite{li2020content}, texture~\cite{yang2020fda}, structure~\cite{tsai2018learning}, and class distribution~\cite{Zou_2018_ECCV}.
To deal with it, many approaches have been proposed, including adversarial training~\cite{tsai2018learning}, entropy minimization~\cite{vu2019advent}, and self-training~\cite{Zou_2018_ECCV, mei2020instance, zhang2021prototypical}.

Among them, cross-domain data mixing based approaches~\cite{chen2021semi,zhou2022context,tranheden2021dacs,gao2021dsp} recently show state-of-the-art performances. 
These methods composite the images of both domains and corresponding (pseudo) labels to generate domain-mixed training samples.
Early works~\cite{chen2021semi} are largely inspired by a popular data augmentation method, CutMix~\cite{yun2019cutmix}, and borrow some rectangular patches from one domain to fill-in the random hole of other domain of image.
Many variants improve data mixing strategy with mixing the region of randomly sampled classes~\cite{tranheden2021dacs}, heuristics on relationship between classes~\cite{zhou2022context}, and image-level soft mixup~\cite{gao2021dsp}.

Motivated by the progress, we further delve into domain mixing approaches and propose \textbf{B}idirectional \textbf{D}omain \textbf{M}ixup (\textbf{BDM}) framework.
Beyond the previous unidirectional sample mixing~\cite{chen2021semi,zhou2022context,tranheden2021dacs,gao2021dsp}, the framework mix the samples in both direction, mix the source patches on the target sample (\textit{i.e.}source-to-target) and vice versa (\textit{i.e.}target-to-source).
Under the framework, we systematically study what makes good domain mixed samples to learn the transferable and generalized features for the target domain performance.
Specifically, we mainly investigate two core steps of data mixing approach: 1) \textbf{Cut}: how can we identify uninformative patches and 2) \textbf{Paste}: which patches from other domains bring better supervision signals.

First, we promote to learn domain transferable and generalized features by cutting out the source-specific and nosily predicted region for source and target data, respectively.
We found that the regions with low-confident model predictions is highly related to it, and cutting out such regions.
To supplement scarce supervisory signal due to the cut process, we design the paste step to fulfill the three key functionalities:
1) As semantic segmentation network heavily rely on the context~\cite{choi2020cars,chen2017deeplab,zheng2021rethinking,yi2022using}, it is important to \textbf{maintain intrinsic spatial structure} of images. Thus we leverage spatial continuity to pick a patch that will be pasted on given the hole region.
2) Previous data mixing approaches~\cite{zhou2022context,tranheden2021dacs} usually paste the randomly selected classes with its correlated classes.
This design choice exacerbates the class imbalanced problem~\cite{gupta2019lvis}, resulting in low performance in sample-scarce class.
Instead, we induce \textbf{class-balanced learning} by giving the high probability to paste the patches with rare classes.
3) Lastly, proposed mixing method \textbf{prevent the noisy learning} by avoiding to paste low-confident patches.

We combine these findings to achieve a new state-of-the-art in the standard benchmarks of domain adaptive semantic segmentation, GTA5 $\rightarrow$ Cityscapes and SYNTHIA $\rightarrow$ Cityscapes setting.
In addition, BMD consistently provides significant performance improvements when build upon the various representative UDA approaches, including adversarial training~\cite{tsai2018learning}, entropy minimization~\cite{vu2019advent}, and self-training~\cite{mei2020instance, zhang2021prototypical}.

\section{Related Works}
\paragraph{Unsupervised Domain Adaptive Semantic Segmentation.}
Unsupervised domain adaptation (UDA) aims to transfer the knowledge learned from the label-rich source domain to the unlabeled target domain.
Early deep-learning-based approaches can be largely categorized into two groups.
\textit{Adversarial-based methods} are to align the feature distribution between the source and target domain via adversarial learning.
Toward this goal, previous works investigate the three different levels: image level~\cite{murez2018image,gong2019dlow,park2019preserving,chen2019crdoco}, intermediate feature level~\cite{hoffman2018cycada,liu2021source,isobe2021multi}, and output level~\cite{tsai2018learning,li2019bidirectional,park2020discover}.
\textit{Self-training-based methods} retrain the models on unlabeled target data with pseudo labels, which are generated based on the model's predictions.
Many works put research efforts into developing the advanced strategies for pseudo-label generation, such as image adaptive thresholding~\cite{mei2020instance} and online pseudo-label generation~\cite{zhang2021prototypical,araslanov2021self}.
Recently, some works~\cite{yang2020label,wang2020classes,li2022class} combine both types of approaches to achieve state-of-the-art performances.

\paragraph{Cross-domain Data Mixing Based Approaches.}
The cross-domain data mixing-based approach is one of the recent breakthroughs in domain adaptative semantic segmentation. 
These methods mix images from the different domains and utilize the mixed samples to train the generalized models.
As an early attempt, DACS~\cite{tranheden2021dacs} borrows a popular data augmentation~\cite{olsson2021classmix} to mix the images, where the source patches of random classes are augmented to target images.
Follow-up works devise additional regularization loss between original and mixed images by proposing regional contrastive consistency~\cite{zhou2022domain} or focusing on boundary regions~\cite{liu2021bapa}.
As another line, there are some trials to present new data mixing strategies.
DSP~\cite{gao2021dsp} softly mixes the images from the source and target domain and uses the corresponding ratio to balance the loss of each domain.
CAMix~\cite{zhou2022context} considers contextual relationships in data mixing by relying on a prior class relationship.

Apart from most previous works, our framework generates the domain-mixed samples in a bidirectional way, source-to-target and target-to-source.
Here we thoroughly study which factors make an effective domain-mixed sample in the context of domain-adaptive semantic segmentation.

\begin{figure*}[!t]
  \includegraphics[width=\linewidth]{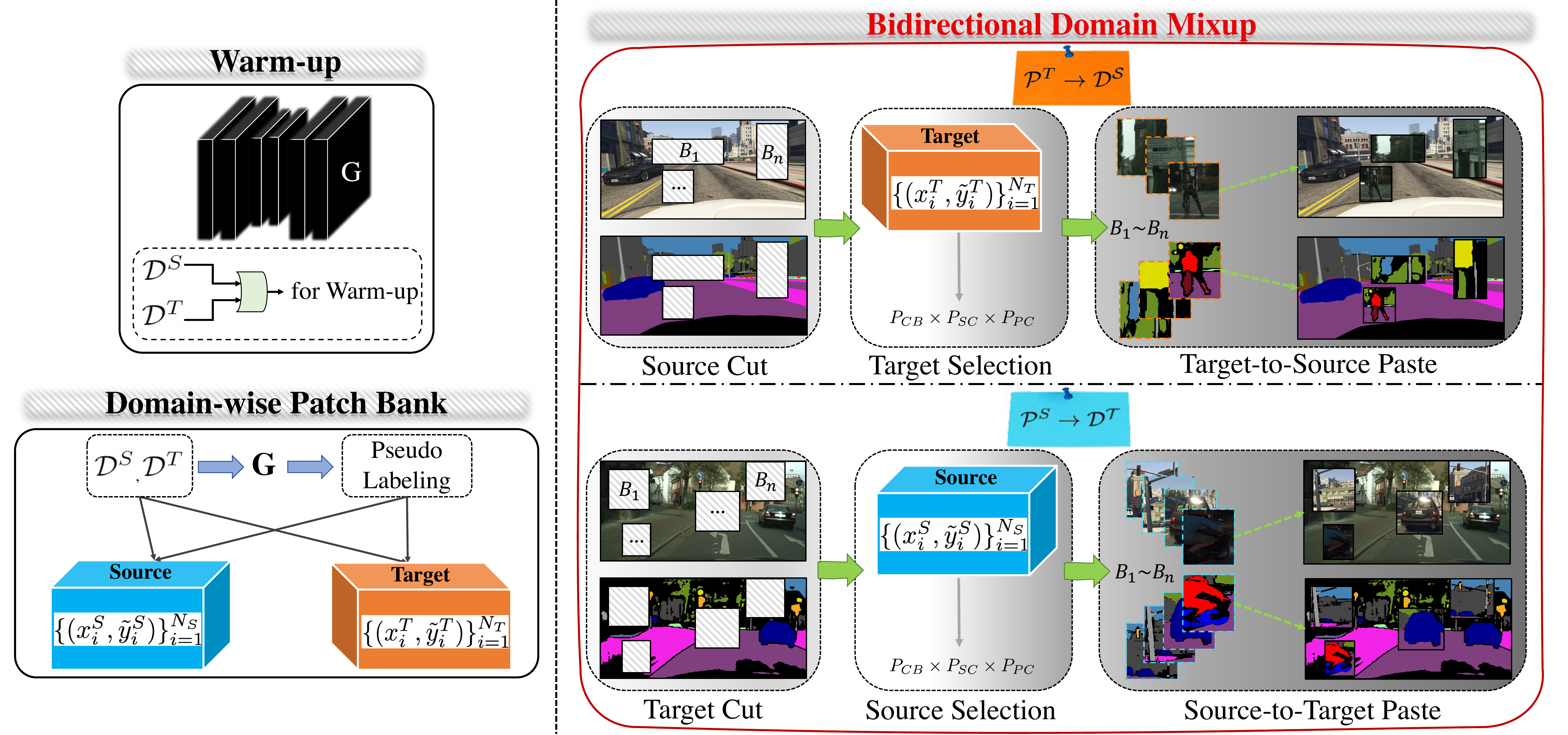}
  \caption{\textbf{Overview of BDM framework.}
  Before training a network with the proposed BDM framework, we first warm up the model with any previous UDA method.
  Given the model $G(.)$, we generate pseudo labels for the source and target domain and store the images and corresponding pseudo labels in the domain-wise patch bank.
  During the training, the framework \textbf{cut} the patches of one domain sample and \textbf{paste} the selected patches from the cross-domain patch bank.
  The mixup process is conducted in a bidirectional way, target-to-source $P_{T}\rightarrow P_{S}$ and source-to-target $P_{S}\rightarrow P_{T}$.
  These samples are guided models to learn domain generalized features.
 }
\label{fig:overview}
\end{figure*}

\section{Method}
This section presents a bidirectional domain mixup framework for domain adaptive semantic segmentation.
To bridge source and target domains that come from different distributions, we simulate intermediate domains by generating domain mixed samples.
{In the next section}, we first describe the overall pipeline to train a network with domain-mixed samples.
Next, we introduce how these samples are generated in detail.

\subsection{The Bidirectional Domain Mixup Framework}
\label{sec:overview}

Given a labeled source dataset $\mathcal{D}^{S} = \{(x_{i}^{S}, y_{i}^{S})\}_{i=1}^{N_{S}}$, and a unlabeled target dataset $\mathcal{D}^{T} = \{x_{i}^{T}\}_{i=1}^{N_{T}}$, our goal is to transfer the knowledge learned from source domain to unlabeled target domain.
To do so, we utilize domain-mixed samples and propose a new data mixing method, bidirectional domain mixup (BDM), to generate them.
As illustrated in ~\figref{fig:overview} and ~\algoref{alg:framework}, this framework comprises the following four major components.

\begin{itemize}[leftmargin=*]
    \item We first apply a previous domain adaptation method as a ~\textit{warm-up}. Our framework allows any previous domain adaptation method. To show the generality, we choose the representative methods including adversarial training~\cite{tsai2018learning}, entropy minimization~\cite{vu2019advent}, and self-training~\cite{mei2020instance, zhang2021prototypical}.   
    \item \textit{Domain-wise patch banks}, $B^{S}$ and $B^{T}$, are constructed to store the samples of each domains.
    We divide images and corresponding pseudo labels into non-overlapping patches and the resulting pairs are stored in domain-wise patch banks.
    For the both domain, we adopt simple strategy~\cite{yang2020fda} to generate pseudo labels, $\{\tilde{y}_{i}^{S}\}_{i=1}^{N_{S}}$ and $\{\tilde{y}_{i}^{T}\}_{i=1}^{N_{T}}$.
    \item During the training, a minibatch of source and target images, $x^{S}$ and $x^{T}$, are sampled.
    ~\textit{Bidirectional domain mixup(BDM)} generate domain-mixed samples, $x_{mix}^{S}$ and $x_{mix}^{T}$, via mixture of images of one domain with patches from other domain.
    Thus, this cross-domain mixing is in bidirectional way, source-to-target and target-to-source.
    Specifically, some rectangular regions(\textit{i.e.} patches) in target samples are cut and source patches retrieved from the source patch bank are pasted (\textit{i.e.} source-to-target direction), and vice versa.
    We also apply same operation on labels, resulting in domain-mixed labels, $y_{mix}^{S}$ and $y_{mix}^{T}$.
    \item Given the mixed images and labels, a segmentation network is trained with standard cross-entropy losses $L_{cross}$. The final loss is formulated as follows: $L_{final} = L_{cross}(x_{mix}^{S}, y_{mix}^{S}) + L_{cross}(x_{mix}^{T}, y_{mix}^{T})$.
    
\end{itemize}

With the framework, we systematically study what makes good domain mixed samples to narrow the domain gap.
Specifically, we mainly investigate two aspects: 1) how can we identify uninformative patches and 2) which patches from other domains bring good supervision signals for the target performance.

\begin{algorithm}[htbp]

\caption{Bidirectional Domain Mixup}\label{algorithm}
\textbf{Input: }warmup model G, labeled source $\mathcal{D}^{S}=\{(x_{i}^{S},y_{i}^{S})\}_{i=1}^{N_{S}}$, unlabeled target $\mathcal{D}^{T}=\{{x_{i}^{T}}\}_{i=1}^{N_{T}}$, cutout $\gamma$ in \eqqref{eq:cutout_module}, horizontal W, Vertical H in {\eqqref{eq:2}}\\

\begin{algorithmic}[1]

\FOR{\texttt{i} = 1 \textbf{to} $N_{S}$, $N_{T}$}
    \STATE
    $\tilde{y}_{i}^{S}$, $\tilde{y}_{i}^{T}$ = Pseudo(G($x_{i}^{S}$)), Pseudo(G($x_{i}^{T}$))
    \STATE
    $B^{S}$, $B^{T}$ = Divide(\{$x_{i}^{S}$, $\tilde{y}_{i}^{S}$\}, \{$x_{i}^{T}$, $\tilde{y}_{i}^{T}$\}, W, H)
\ENDFOR{}

    \STATE
    $\{\hat{x}_{i}^{S}, \hat{y}_{i}^{S}\} = \text{Cut}(\{x_{i}^{S}, \tilde{y}_{i}^{S}\}, \gamma)$ 
    \STATE
    $\mathcal{P}_{selected}^{T} = \text{Selection}(B^{T}(P_{CB}, P_{SC}, P_{PC}))$
    \STATE
    $\{{x}_{mix}^{S}, {y}_{mix}^{S}\} = \text{Paste}(\{\hat{x}_{i}^{S}, \hat{y}_{i}^{S}\}, \mathcal{P}_{selected}^{T})$ 
    \STATE
    $\{\hat{x}_{i}^{T}, \hat{y}_{i}^{T}\} = \text{Cut}(\{x_{i}^{T}, \tilde{y}_{i}^{T}\}, \gamma)$
    \STATE
    $\mathcal{P}_{selected}^{S} = \text{Selection}(B^{S}(P_{CB}, {P}_{SC}, P_{PC}))$
    \STATE
    $\{{x}_{mix}^{T}, {y}_{mix}^{T}\} = \text{Paste}(\{\hat{x}_{i}^{T}, \hat{y}_{i}^{T}\}, \mathcal{P}_{selected}^{S})$ 
    \STATE
    \textbf{return} $\{{x}_{mix}^{S}, {y}_{mix}^{S}\}, \{{x}_{mix}^{T}, {y}_{mix}^{T}\}$ 
    
\end{algorithmic}
\label{alg:framework}
\end{algorithm}

\subsection{Cut}
\label{sec:cut}

Cut is a process that masks out contiguous sections (\textit{i.e.} multiple rectangular regions) of input and corresponding labels.
We introduce widely used random patch cutout~\cite{devries2017improved,yun2019cutmix,tranheden2021dacs} and proposed confidence based cutout.
\paragraph{Random region cutout.} Random region cutout is a simple but strong baseline that is adopted in various tasks such as UDA~\cite{tranheden2021dacs}, Semi-DA~\cite{chen2021semi} and SSL~\cite{wang2022semi}.
When $x$ and $y$ are input, random region cutout could be formulated as follows:
\begin{equation}
  x_{cut} = M \odot x, y_{cut} = M \odot y,
  \label{eq:cutout_random}
\end{equation}

where $\mathbf{M} \in \{0,1\}^{W \times H}$ denotes a binary mask indicating where to drop out, and $\odot$ is element-wise multiplication. 
Here, we zero out the multiple random regions of the mask to cut. 
\paragraph{Confidence based cutout.} 
The random region cut the regions regardless of whether it is informative.
Instead, we target to cut where provide noisy supervision in terms of learning the generalized features.
To this end, we see that it is important to discard the regions with low confident predictions for the following reasons:
1) for the source domain, it remove non-transferable and source specific regions {(please refer to discussion section.)},
2) it prevent to learn from noisy pseudo labels of target data.

Given the source images $x_{i}^{S}$ and their pseudo labels $\tilde{y}_{i}^{S}$, we calculates the ratio of the uncertain region over the randomly generated region $\mathbf{B}= (r_{x},r_{y},r_{w},r_{h})$.
Here we see the ignored region of pseudo labels as the uncertain region.
If the ratio of the uncertain region is above the cutout threshold $\gamma$, the region is cut.
The proposed cutout is summarized as follows:

\begin{equation}
  \hat{x}_{i}^{S}, \hat{y}_{i}^{S} =
  \begin{cases}
  \textrm{Cutout($x_{i}^{S}, \tilde{y}_{i}^{S}$)}, & \textrm{if}\quad \mathcal{H}(\tilde{y}_{i}^{S},\mathbf{B})> \gamma, \\
  x_{i}^{S}, \tilde{y}_{i}^{S} & \textrm{otherwise},
  \end{cases}
  \label{eq:cutout_module}
\end{equation}
where $\mathcal{H}$ denote a function that computes the ratio of the uncertain region over region $\mathbf{B}$.
And, the threshold $\gamma$ is set to 0.2.
Similar to ~\eqqref{eq:cutout_module}, low confident regions of target samples are also masked.

\subsection{Paste}
\label{sec:paste}

To generate the domain-mixed samples, we sample the patch from the other domain and paste it back to the region that are cutout.
This enables the resulting samples to include both source and target patches.
While already effective (see~\tableref{table:ablation_3}), we see that there are more room for the improvement.

We additionally consider three important factors during the patch sampling.
First, we introduce \textbf{class-balanced} sampling.
As random sampling tend to bias the mixup samples mainly toward the frequent classes, we offset this undesirable effect using class-balanced sampling.
Since the long-tailed category distribution tends to be similar for different domains, we compute the class distribution of the source domain~\cite{zhang2021prototypical, li2022class} to provide more chance for patches that include rare classes.
Second, we consider \textbf{spatial continuity} of the natural scene.
The random sampling can produce mixed samples that are geometrically unnatural, and thus causes severe train and test time inconsistency.
Instead, based on the fact that the vehicle egocentric view has strong fixed spatial priors~\cite{choi2020cars,zou2018domain}, we compute spatial priors for each semantic class~\cite{zou2018domain} in the source domain and use this statistics during sampling. In this way, we successfully synthesize geometrically reasonable mixed samples (see~\figref{fig:total_vis}).
Finally, we take \textbf{pseudo label confidence} into account.
The rationale behind this strategy is to increase the sampling probability of the patches that include confident pixels (and vice versa).
This is also strongly connected to the cut operation, where we dropped the regions that are not domain transferable, while here we perform the opposite.
We observe that all these components impact the final performance, and the best mixup results are achieved when these are used jointly.

\subsubsection{Patch Generation.}
\label{sec:patch_generation}

In the source and target datasets, classes such as \textit{train} and \textit{bike} have a small number of samples, so to consider the class online, it is necessary to find a sample including a 6 rare class from all samples.
Since this method is memory inefficient, we cut the patch containing each class and save it.

We generate a patch by cutting the pseudo labeled datasets $\{(x_{i}^{S}, \tilde{y}_{i}^{S})\}_{i=1}^{N_{S}}$ and $\{(x_{i}^{T}, \tilde{y}_{i}^{T})\}_{i=1}^{N_{T}}$ at equal intervals by the number of horizontal $\mathbf{W}$ and vertical $\mathbf{H}$.
Therefore, one image and its pseudo label are divided into $\mathbf{W} \times \mathbf{H}$ patches.
After that, the patches extracted from each location are grouped into one patch sequence.
Next, the patch containing the class $\{C_{i}\}_{i=1}^{K}$, $\mathbf{K}$ is the number of classes, from the pseudo label of the patch sequence representing each spatial location is stored in the child patch sequence, and it is possible to duplicate it.
Therefore, patches considering spatial location and class existence are grouped into a total of $\mathbf{W} \times \mathbf{H} \times \mathbf{K}$ patch sequence.
Finally, at the patch sequence of $\mathbf{W} \times \mathbf{H} \times \mathbf{K}$, the normalized confidence calculated by the method in {~\eqqref{eq:3}} is sorted in ascending order, and $\mathbf{R}$ patch sequences are generated at regular intervals.
The number of finally generated patch sequence is $\mathbf{W} \times \mathbf{H} \times \mathbf{K} \times \mathbf{R}$

\subsubsection{Patch Selection}
\label{sec:patch_selection}

\paragraph{Class-balanced Patch Sampling.}
In addition to Zipfian distribution of object categories~\cite{manning1999foundations}, 
the difference in the intrinsic size of each object makes pixel-level class imbalances more severe in semantic segmentation.
However, the random patch sampling for paste make bias toward the frequent classes, as it naturally follow the probability of existence.

To alleviate the class imbalance problem in paste process, we propose patch-level oversampling.
Given the total number of pixels for each classes in the source labels $\{\bar{N}^{i}\}_{i=1}^{K}$, the probability $P_{CB}$ that each class patch is selected can be formulated as follows:
\begin{equation}
\begin{aligned}
  \hat{P}_{CB} &= \{(-\log(\frac{\bar{N}^{i}}{\sum_{i=0}^{\mathbf{K}}{\bar{N}^{i}}}))^{\alpha}\}_{i=1}^{K},\\
  \\
  P_{CB} &= \{(\frac{\hat{P}_{CB}^{i}}{\sum_{i=0}^{\mathbf{K}}{\hat{P}_{CB}^{i}}})\}_{i=1}^{K},
  \label{eq:1}
\end{aligned}
\end{equation}
where $\mathbf{K}$ is the number of classes, $\alpha$ is the sharpening coefficient. We set $\alpha$ to 2 in all experiments.

\paragraph{Sampling with Spatial Continuity.} 
The spatial layout between the source (synthetic) and target dataset share large similarities.
For example, there are common rules like cars can't fly in the sky ~\cite{choi2020cars} in both the domains.
Therefore, we propose spatial continuity based paste that considers spatial relationship instead of pasting patches at random positions.
The probability of selecting each location $\{\textrm{Patches}_{i}\}_{i=1}^{\mathbf{W} \times \mathbf{H}}$ of the patches generated through {Patch Generation section} to be mixed with the patch locations cutout through {Cut section} is calculated as follows:

\begin{equation}
\begin{aligned}
    \hat{SC} &= \textrm{argmax}\{\textrm{SC}_{i}(o_{w},o_{h})_{i=1}^{K}\}, \\
    P_{SC} &= \{\hat{SC}(\textrm{Patches}_{i}(o_{\hat{w}^{i}},o_{\hat{h}^{i}})) \}_{i=1}^{\mathbf{W}\times \mathbf{H}},
  \label{eq:2}
\end{aligned}
\end{equation}
where $\{\textrm{SC}_{i}\}_{i=1}^{\mathbf{K}}$ is the source domain class-wise spatial prior kernel map generated by CBST-SP~\cite{zou2018domain}. where $o_{h},o_{w}$ is the center coordinates of the cutout patch, $o_{\hat{h}}^{i},o_{\hat{w}}^{i}$ is the center coordinates of the patch at the i-th location.
Note that, we normalized the sum of the set $P_{SC}$ to 1.

\begin{table*}[t]
\resizebox{1.\textwidth}{!}{
\begin{threeparttable}
 \centering
 \def\arraystretch{1.1}
 \begin{tabular}{c|c|c|ccccccccccccc|cccccc}
    \hline \hline
    & & &\multicolumn{13}{c}{Head Classes}&\multicolumn{6}{|c}{Tail Classes}\\
    \hline
    Method  &\rotatebox{90}{mIoU} &\rotatebox{90}{mIoU-tail} &{\rotatebox{90}{road}} & {\rotatebox{90}{sidewalk}} & {\rotatebox{90}{building}} & {\rotatebox{90}{wall}} & {\rotatebox{90}{fence}} & {\rotatebox{90}{pole}} & {\rotatebox{90}{vegetation}} & {\rotatebox{90}{terrain}} & {\rotatebox{90}{sky}} & {\rotatebox{90}{person}} & {\rotatebox{90}{car}} & {\rotatebox{90}{truck}} & {\rotatebox{90}{bus}} & {\rotatebox{90}{light}} & {\rotatebox{90}{sign}} & {\rotatebox{90}{rider}} & {\rotatebox{90}{train}} & {\rotatebox{90}{motorcycle}} & {\rotatebox{90}{bike}}\\
    \hline
    Source Only & 36.6&24.0&75.8&16.8&77.2&12.5&21.0&25.5&81.3&24.6&70.3&53.8&49.9&17.2&25.9&30.1&20.1&26.4&6.5&25.3&36.0\\
    \hline
    Adaptseg~\cite{tsai2018learning} &  41.4&25.0&86.5&25.9&79.8&22.1&20.0&23.6&81.8&25.9&75.9&57.3&76.3&29.8&32.1&33.1&21.8&26.2&7.2&29.5&32.5\\
    ADVENT~\cite{vu2019advent}  &  45.5&25.7&89.4&33.1&81.0&26.6&26.8&27.2&83.9&36.7&78.8&58.7&84.8&38.5&44.5&33.5&24.7&30.5&1.7&31.6&32.4\\
    CCM~\cite{li2020content} & 49.9&26.9&93.5&57.6&84.6&39.3&24.1&25.2&85.0&40.6&86.5&58.7&85.8&49.0&56.4&35.0&17.3&28.7&5.4&31.9&43.2\\
    IAST~\cite{mei2020instance} & 51.5&34.0&93.8  &57.8&85.1&39.5&26.7&26.2&84.9&32.9&88.0&62.6&87.3&39.2&49.6&43.1&34.7&29.0&23.2&34.7&39.6\\
    DACS~\cite{tranheden2021dacs} &  52.1&32.6&89.9&39.6&87.8&30.7&39.5&38.5&87.9&43.9&88.7&67.2&84.4&45.7&50.1&46.4&52.7&35.7&0.0&27.2&33.9\\

    DSP~\cite{gao2021dsp} &  55.0&36.9&92.4&48.0&87.4&33.4&35.1&36.4&87.7&43.2&89.8&66.6&89.9&57.1&56.1&41.6&46.0&32.1&0.0&44.1&57.8\\

    CAMix~\cite{zhou2022context} & 55.2&37.9&93.3&58.2&86.5&36.8&31.5&36.4&87.2&44.6&88.1&65.0&89.7&46.9&56.8&35.0&43.5&24.7&27.5&41.1&56.0\\
    CorDA~\cite{wang2021domain}& 56.6 & 38.5 & \textbf{94.7} & {63.1} & 87.6& 30.7 & 40.6 & 40.2 & 87.6 & 47.0 & \textbf{89.7} & 66.7 & 90.2 & 48.9 & 57.5 & 47.8 & 51.6 & 35.9 & 0.0 & 39.7 & 56.0\\
    ProDA~\cite{zhang2021prototypical} & 57.5&42.0&87.8&56.0&79.7&46.3&44.8&45.6&88.6&45.2&82.1&70.7&88.8&45.5&59.4&53.5&53.5&39.2&1.0&48.9&56.4\\
    DAP~\cite{huo2022domain}& 59.8 & 44.6 & 94.5 & \textbf{63.1} & 89.1& 29.8 & 47.5 & 50.4 & 89.5 & 50.2 & 87.0 & \textbf{73.6} & \textbf{91.3} & 50.2 & 52.9 & 56.7 & 58.7 & 38.6 & 0.0 & 50.2 & \textbf{63.5}\\

    \hline

    Adaptseg~\cite{tsai2018learning} + Ours &  57.4\textcolor{blue}{(+16.0)}&44.4&89.3&50.0&88.4&45.6&45.4&41.1&78.0&35.6&82.3&69.2&87.5&\textbf{55.7}&57.8&49.9&\textbf{60.2}&45.6&8.1&45.5&57.2\\
    ADVENT~\cite{vu2019advent} + Ours & 57.6\textcolor{blue}{(+12.1)}&38.9&91.3&51.8&86.7&49.9&49.2&53.3&85.8&47.9&85.7&62.3&87.8&55.5&54.4&43.1&43.3&45.9&4.4&46.3&50.4\\
    IAST~\cite{mei2020instance} + Ours & 61.0\textcolor{blue}{(+9.5)}&46.5&92.1&59.6&\textbf{89.9}&52.9&55.7&49.2&89.3&46.7&86.3&59.1&88.3&54.8&55.9&44.6&45.8&42.0&\textbf{39.2}&50.3&57.3\\
    ProDA~\cite{zhang2021prototypical} + Ours &  \textbf{63.9}\textcolor{blue}{(+6.4)}&\textbf{47.8}&89.2&60.1&83.8&\textbf{61.5}&\textbf{63.6}&\textbf{66.7}&\textbf{90.4}&\textbf{51.1}&83.5&{72.6}&88.0&51.2&\textbf{65.3}&\textbf{58.2}&59.3&\textbf{47.8}&1.0&\textbf{60.1}&{60.9}\\
    
    \hline
    
    Target Only  &64.5&53.0&96.2&75.5&87.7&38.0&39.6&43.4&88.2&52.4&89.5&69.7&91.4&66.2&69.7&46.6&62.8&49.5&45.0&49.0&65.1 \\
    
    \hline
    \hline
    \end{tabular}
    \end{threeparttable}    
    }
    \caption{Comparison with state-of-the-art models on GTA5 $\rightarrow$ Cityscapes. We highlight the mIoU of tail classes (\textit{i.e.} mIoU-tail) along with per-class IoU and overall mIoU.
    Our results are averaged over five runs.}
    \label{table:gta5}
\end{table*}

\paragraph{Sampling with Normalized Confidence.}
\label{sec:GC}

Opposite to confidence based cutout, in the paste, we give high probability to the patches that include confident pixels.
To faithfully measure the confidence level of a patch, we take the difficulty of each class into account and design the normalized confidence of a patch.
We first calculate the average confidence of classes using the set of pseudo labels and use it to represent the difficulty of each class.
Then, the confidence of patches in the patch bank is normalized at pixel-level by subtracting the difficulty score according to predicted classes (Norm).
Intuitively, it measure the \textit{relative} confidence level.
The resulting normalized confidence maps are spatially averaged (Average), and patches are sorted in ascending order according to it (Sort).

\begin{equation}
\begin{aligned}
    \hat{B}^{T} &= \{Average(Norm(B_{i}^{T}))\}_{i=1}^{N_{T}}, \\
    \hat{B}^{S} &= \{Average(Norm(B_{i}^{S}))\}_{i=1}^{N_{S}}, \\
    \bar{B}^{T} &= Sort(\hat{B}^{T}), \\
    \bar{B}^{S} &= Sort(\hat{B}^{S})
  \label{eq:3}
\end{aligned}
\end{equation}

Finally, the patch is divided into three batches: the low, the middle, and the high confidence group.
The probability $P_{PC}$ that the patch in each group is selected is $\{0.1, 0.3, 0.6\}$.

\paragraph{Probability of selection of each patch sequence.}
We select the patch jointly considering class balance, spatial continuity, and pseudo label confidence for BDM.
Since each probability is independent, the probability that each patch sequence is selected is $P_{CB} \times P_{SC} \times P_{PC}$.

\section{Experiments}
In this section, we present experimental results to validate the proposed BDM for domain adaptive semantic segmentation.

We first describe experimental configurations in detail.
After that, we validate our BDM on two public benchmark datasets, GTA5 and SYNTHIA~\cite{ros2016synthia} datasets, and provide detailed analyses.
Note that the Intersection-over-Union (IoU) metric is used for all the experiments.

\subsection{Experimental Settings}
\paragraph{Dataset.}
We evaluate our proposed Bidirectional Domain Mixup on two popular domain adaptive semantic segmentation benchmarks(SYNTHIA $\rightarrow$ Cityscapes, and GTA5 $\rightarrow$ Cityscapes).
Cityscapes~\cite{cordts2016cityscapes} is a real-world urban scene dataset consisting of a training set with 2,975 images, a validation set with 500 images and a testing set with 1,525 images. We use the unlabeled training dataset as $\{D_{i}^{T}\}_{i=1}^{2,975}$ and evaluate our Bidirectional Domain Mixup with 500 images from the validation set.
SYNTHIA~\cite{ros2016synthia} is a synthetic urban scene dataset. We pick SYNTHIA-RAND-CITYSCAPES subset as the source domain, which shares 16 semantic classes with Cityscapes. In total, 9,400 images from SYNTHIA dataset are used as source domain training data $\{D_{i}^{S}\}_{i=1}^{9,400}$ for the task.
GTA5~\cite{richter2016playing} dataset is another synthetic dataset sharing 19 semantic classes with Cityscapes. 24,966 urban scene images are collected from a physically-based rendered video game Grand Theft Auto V (GTAV) and are used as source training data $\{D_{i}^{S}\}_{i=1}^{24,966}$. 
We view 6 and 5 with relatively few training samples as tail-classes for each source domain(GTA5, SYNTHIA), respectively.

\begin{table*}[t]

\resizebox{1.\textwidth}{!}{
 \centering
 \begin{threeparttable}
 \def\arraystretch{1.1}
 \begin{tabular}{c|c|c|cccccccc|ccccc}
    \hline \hline
    & & &\multicolumn{8}{c}{Head Classes}&\multicolumn{5}{|c}{Tail Classes}\\
    \hline
    Method  & {\rotatebox{90}{mIoU}}& {\rotatebox{90}{mIoU-tail}}&{\rotatebox{90}{road}} & {\rotatebox{90}{sidewalk}} & {\rotatebox{90}{building}}  & {\rotatebox{90}{vegetation}} & {\rotatebox{90}{sky}} & {\rotatebox{90}{person}}  & {\rotatebox{90}{car}} & {\rotatebox{90}{bus}}& {\rotatebox{90}{light}}& {\rotatebox{90}{sign}} & {\rotatebox{90}{rider}} & {\rotatebox{90}{motorcycle}} & {\rotatebox{90}{bike}}\\
    \hline
    Source Only & 40.3&20.8&64.3&21.3&73.1&63.1&67.6&42.2&73.1&15.3&7.0&27.7&19.9&10.5&38.9 \\
    \hline
    Adaptseg~\cite{tsai2018learning} &  45.8&18.5&79.5&37.1&78.2&78.0&80.3&53.7&67.1&29.4&9.3&10.6&19.2&21.8&31.6   \\

    ADVENT~\cite{li2020content} &  48.0&16.9&85.6&42.2&79.7&80.4&84.1&57.9&73.3&36.4&5.4&8.1&23.8&14.2&33.0   \\

    CCM~\cite{li2020content} &  52.9&29.0&79.6&36.4&80.6&81.8&77.4&56.8&80.7&45.2&22.4&14.9&25.9&29.9&52.0   \\

    IAST~\cite{mei2020instance} &  57.0&35.2&81.9&41.5&83.3&83.4&85.0&65.5&86.5&38.2&30.9&28.8&30.8&33.1&52.7 \\

    DACS~\cite{tranheden2021dacs} &  54.8&32.1&80.5&25.1&81.9&83.6&\textbf{90.7}&67.6&82.9&38.9&22.6&23.9&38.3&28.4&47.5  \\

    DSP~\cite{gao2021dsp} &  59.9&35.9&86.4&42.0&82.0&87.2&88.5&64.1&83.8&65.4&31.6&33.2&31.9&28.8&54.0  \\
    
    CAMix~\cite{zhou2022context} &  59.7&33.8&91.8&54.9&83.6&83.8&87.1&65.0&85.5&55.1&23.0&29.0&26.4&36.8&54.1  \\
    CorDA~\cite{wang2021domain}& 62.8 & 41.5 &\textbf{93.3} & \textbf{61.6} & 85.3 & 84.9 & 90.4 & 69.7 & 85.6 & 38.4 & 36.6 & 42.8 & \textbf{41.8} & 32.6 & 53.9\\

    ProDA~\cite{zhang2021prototypical} &  62.0&40.3&87.8&45.7&84.6&{88.1}&84.4&74.2&88.2&51.1&54.6&37.0&24.3&40.5&45.6  \\

    DAP~\cite{huo2022domain}& 64.3 & \textbf{48.7}& 84.2 & 46.5 & 82.5 & \textbf{89.3} & 87.5 & 75.7 & \textbf{91.7} & \textbf{73.5} & 53.6 & \textbf{45.7} & 34.6 & \textbf{49.4} & \textbf{60.5}\\

    \hline

    Adaptseg\tnote{*}~\cite{tsai2018learning} + Ours &  50.6\textcolor{blue}{(+4.8)}&24.1&84.5&44.3&79.5&84.2&83.3&60.1&69.3&33.2&19.6&18.7&23.6&24.0&34.7 \\

    ADVENT~\cite{tsai2018learning} + Ours &  53.8\textcolor{blue}{(+5.8)}&27.1&85.4&47.7&82.9&86.5&85.7&64.5&72.0&39.3&25.2&22.2&25.6&26.1&36.4 \\

    IAST~\cite{mei2020instance} + Ours &  62.9\textcolor{blue}{(+5.9)}&42.6&88.2&50.2&\textbf{88.5}&85.6&89.7&70.2&{88.3}&44.8&42.3&33.5&39.0&39.4&{59.0}    \\

    ProDA~\cite{zhang2021prototypical} + Ours &  \textbf{66.8}\textcolor{blue}{(+4.8)}&{47.7}&91.0&\textbf{55.8}&86.9&85.8&85.7&\textbf{84.1}&86.0&{55.2}&\textbf{58.3}&{44.7}&{40.3}&{45.0}&50.6 \\

    \hline
    Target Only  & 72.3&54.6&96.2&75.5&87.7&88.2&89.5&69.7&91.4&69.7&46.6&62.8&49.5&49.0&65.1 \\

    \hline
    \hline
    \end{tabular}
    \begin{tablenotes}
    \item[*] Note that pre-trained weights are not provided, so use them after reproduction.
    \end{tablenotes}
    \end{threeparttable}
}
\caption{Comparison with state-of-the-art models on SYNTHIA $\rightarrow$ Cityscapes. Our results are averaged over five runs.}
\label{table:SYNTHIA}
\end{table*}

\paragraph{Training.} In all our experiments, we used Deeplabv2~\cite{chen2017deeplab} architecture with pre-trained ResNet101~\cite{he2016deep} as backbone on {ImageNet~\cite{deng2009imagenet} and MSCOCO~\cite{lin2014microsoft}}, and for fair performance evaluation, all hyperparameter settings such as batch size, learning rate, and iteration follow the standard protocol~\cite{tsai2018learning}. 

To show the genearlity of BDM, we choose multiple representative UDA approaches as warm-up method, including Adaptseg~\cite{tsai2018learning} (adversarial training), ADVENT~\cite{vu2019advent} (entropy minimization), IAST~\cite{mei2020instance} (self-training), and ProDA~\cite{zhang2021prototypical} (self-training) as warm-up models, respectively
We use official implementation of them\footnote{https://github.com/wasidennis/AdaptSegNet}\footnote{https://github.com/Raykoooo/IAST}\footnote{https://github.com/microsoft/ProDA}\footnote{https://github.com/valeoai/ADVENT}. Note that all ablation studies were performed with Adaptseg.

Given the warm-up model, we further train the model with proposed BDM framework for 1,000,000 iterations.
For the stable training, EMA framework~\cite{olsson2021classmix} is adopted.
Unless otherwise specified, we set the number of patches $\mathbf{W}$ and $\mathbf{H}$ as 4 and 3.
For the number of randomly generated boxes in cut process, we choose 4 as default.
The number of classes $\mathbf{K}$ is 19 and 13 when we use GTA5 and SYNTHIA as the source domain, respectively.
Finally, the number of pseudo-label reliability intervals, $\mathbf{R}$, was set to 3 in all experiments.

\subsection{Comparison with State-of-the art}

In this section, we compare our proposed method with the top-performing UDA approach.

~\tableref{table:gta5} shows the comparisons on GTA5 $\rightarrow$ Cityscapes setting.
DACS, which classmix at random locations without considering the class distribution of the source dataset, significantly improved performance in classes with a large number of samples, but significantly decreased in tail classes such as \textit{bike} and \textit{train}.
CAMix, a classmix method considering the contextual relationship, solved the problem of performance degradation of tail classes through consistency loss and dynamic threshold. However, considering only the contextual relationship, the performance decreased in \textit{wall}, \textit{light}, and \textit{rider} classes, which have significantly different contextual relationship between GTA5 and Cityscapes.

On the other hand, our BDM jointly consider class balance, spatial continuity, and pseudo label confidence and achieve state-of-the-art with an mIoU score of 63.9\% when ProDA was selected as the warm-up model.
Despite the great overall scores, it showed a low IoU in the \textit{train} class.
The rationale behind this is the severely poor performance of the chosen warm-up model in that class.
Instead, when the warm-up model is switched to IAST, we achieved much improved scores (23.2\% $\rightarrow$ 39.2\%) IoU in the \textit{train} class.

Last but not least, our method shows consistent performance improvement with four different warm-up models, showing the generality of our framework.

~\tableref{table:SYNTHIA} shows the comparisons of SYNTHIA $\rightarrow$ Cityscapes adaptation.
Again, our BDM achieved state-of-the-art with an mIoU score of 66.8\% when ProDA was selected as the warm-up model.
These experimental results indicate that BDM is valid not only in the GTA5 dataset, where the scene layout between the source and target dataset is highly similar but also in the SYNTHIA dataset which includes images with different viewpoints (e.g. top-down view).

\begin{table}[htbp]
\centering
\resizebox{0.45\textwidth}{!}{
 \centering
 \def\arraystretch{1.1}
 \begin{tabular}{l|c|c|c|c|c}
    \hline \hline
    Method & BD. & Cut  & Paste  & mIoU & gain\\ \hline
    Warm-up~\cite{tsai2018learning} & - & - & - & 46.4 & +0 \\ \hline
    \multirow{2}*{DACS~\cite{tranheden2021dacs}} &   -        & Rand  & Rand &  50.9  & +9.5\\ 
                                & \checkmark & Rand  & Rand &  53.6  & +12.2\\ \hline
    \multirow{3}*{Ours} & \checkmark  & Ours & Rand  & 53.8 & +12.4\\
                        & \checkmark  & Rand & Ours  & 56.8 & +15.4\\
                        & \checkmark  & Ours & Ours  & 57.4 & +16.0\\ \hline
    Ours-T2S            & -           & Ours & Ours  & 56.0 & +14.6\\
    Ours-S2T            & -           & Ours & Ours  & 54.8 & +13.4 \\

    \hline
    \hline
    \end{tabular}}
    \caption{Component-wise Ablation Study.}
    \label{table:ablation_1}
\end{table}

\begin{figure*}[t!]
  \centering
  \includegraphics[width=0.8\linewidth]{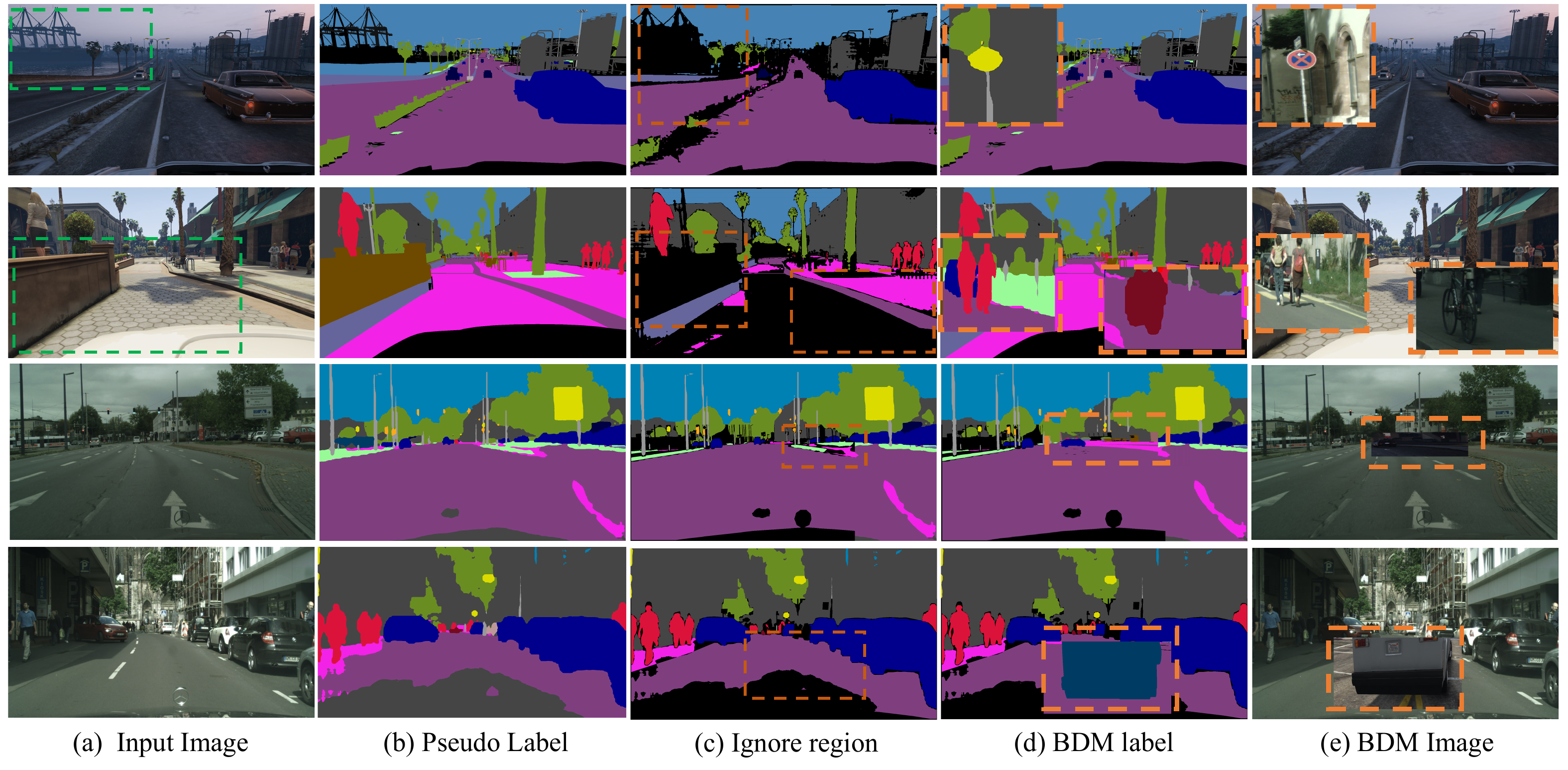}
  \caption{Qualitative results of cross domain mixed samples generated with the proposed BDM. Two rows from top to bottom: target-to-source, source-to-target. green box has a high probability of being Cut because ignore pixels are dense.\vspace{-2.0mm}
}
  \label{fig:total_vis}
\end{figure*}

\subsection{Ablation Study}
\label{sec:ablation}

\paragraph{Component-wise Ablation.}

First, we explore the importance of the bidirectional (\textbf{BD.}) mixup (please refer to \tableref{table:ablation_1}).
We implement two unidirectional variants, Ours-T2S and Ours-S2T, which mix the samples in target-to-source and source-to-target, respectively.
As degraded results indicate, both directions of the mixture are important, showing the effectiveness of our BDM framework.
We also report the results of DACS~\cite{tranheden2021dacs} and its bidirectional extension.
For a fair comparison, we re-implement their mixing strategy on our framework.
Again, the design of bidirectional mixing improves the performance by 2.7 mIoU.

We also investigate the impact of the cut and paste mechanism.
Both proposed mechanism outperforms the popular random baseline.
In particular, when we apply a random baseline for both the cut and paste, the model degenerate into a bidirectional extension of DACS~\cite{tranheden2021dacs}.
Compared to it, our final model boosts the performance by 3.8 mIoU.

\paragraph{Ablation Study on Paste.}
~\tableref{table:ablation_3} study the effects of class balance (CB), spatial continuity (SC), and pseudo label confidence (PC) on the paste mechanism.
Ours$_{cut}$ refers to the model trained with only the proposed cut process.
As shown in ~\tableref{table:ablation_3}, the CB component showed the largest performance improvement.  
Also, SC and PC showed 0.3 mIoU and 0.9 mIoU additional improvement, respectively.
In addition, the best performance was achieved when all three components were used. These experimental results indicate that all components of CB, SC and PC are complementary in BDM.

\begin{table}[htbp]
\centering

\resizebox{0.45\textwidth}{!}{
 \centering
 \def\arraystretch{0.65}
 \begin{tabular}{c|c|c|c|c|c}
    \hline \hline
    \tiny{Method} & \tiny{CB}  & \tiny{SC} & \tiny{PC}  & \tiny{mIoU} & \tiny{gain}\\ \hline
    \tiny{Ours$_{cut}$}         & -  & - &  - &  \tiny{48.9}  & \tiny{+0} \\ \hline
    \multirow{5}*{\tiny{Ours}}  & -  & - &  - &  \tiny{53.8}  & \tiny{+4.9}  \\
                         & \checkmark  & - & -  &\tiny{56.1} & \tiny{+7.2}\\
                         & \checkmark  & \checkmark & -  & \tiny{56.4}& \tiny{+7.5} \\
                         & \checkmark  & - & \checkmark  & \tiny{57.0} & \tiny{+8.1}\\
                         & \checkmark  &  \checkmark& \checkmark  & \tiny{57.4} & \tiny{+8.5}\\
    \hline
    \hline
    \end{tabular}}
    \caption{Ablation study on Paste.}
    \label{table:ablation_3}
\end{table}

\begin{table}[htbp]
\centering

\resizebox{0.40\textwidth}{!}{
 \centering
 \def\arraystretch{1.0}
 \begin{tabular}{c|c|c|c}
    \hline \hline
     & No Cut  & Cut Top 50\% Conf. & Cut Bottom 50\% Conf. \\ \hline
    mIoU & 36.6  & 27.8 \textcolor{red}{(-8.8)} & 41.3 \textcolor{blue}{(+4.7)} \\ \hline

    \hline
    \hline
    \end{tabular}}
    \caption{Analysis on source cut region.}
    \label{table:ablation_2}
\end{table}

\begin{table}[htbp]
\centering

\resizebox{0.35\textwidth}{!}{
 \centering
 \def\arraystretch{0.70}
 \begin{tabular}{c|c|c}
    \hline \hline
    {Model} & {Recall}  &  {Precision} \\ \hline
    {Warm-up}~\cite{tsai2018learning} & {97.6}  & {73.2} \\
    \hline
    \hline
    \end{tabular}}
    \caption{Analysis on Low-confident Regions.\vspace{-3.0mm}}
\label{table:ablation_4}
\end{table}

\section{Discussion}
\label{sec:discussion}

\paragraph{Connection between Low-confident and Transferable Region on Source Domain.} 

We aim to cutout the non-transferable regions of the source domain for learning generalized features.
To realize it, we remove the patches with low confident prediction.
Here, we validate the design by showing that low-confident source regions are highly correlated with non-transferable regions.
To this end, we first train multiple source-only models with varying regions of cut and measure the performance on the target dataset (see \tableref{table:ablation_2}).
Compared to the default model that uses all the source labels (\textit{i.e.} No Cut in the \tableref{table:ablation_2}), the performance is notably increased by 4.7 mIoU when we discard the labels of the bottom 50\% of low confident pixels.
On the contrary, cutting the high confident region makes significant performance degradation.
It implies that training models with low confident pixels promote overfitting to the source domain.

Moreover, we examine how the selected low-confident regions correlates with lower domain transferrablity. 
Based on the assumption that the target-only model has oracle feature representations, we compare the overlapping regions between the 
selected low-confident regions from our strategy and the target-only model. The results are summarized in~\tableref{table:ablation_4}.
We see the regions highly overlap, implying that the simple threshold-based cutout can already leave the highly transferrable regions, and help better adaptation of the model.

\paragraph{Qualitative analysis of Bidirectional Domain Mixup.}
~\figref{fig:total_vis} is a sample of training images generated through BDM.
BDM receives ~\figref{fig:total_vis}-(c) and cuts the ignore region, and then pastes another patch to the corresponding region as shown in  ~\figref{fig:total_vis}-(d).
Our framework is trained with ~\figref{fig:total_vis}-(e) and ~\figref{fig:total_vis}-(d) as pairs.

\begin{figure}[h!]
  \centering
  \includegraphics[width=1.0\linewidth]{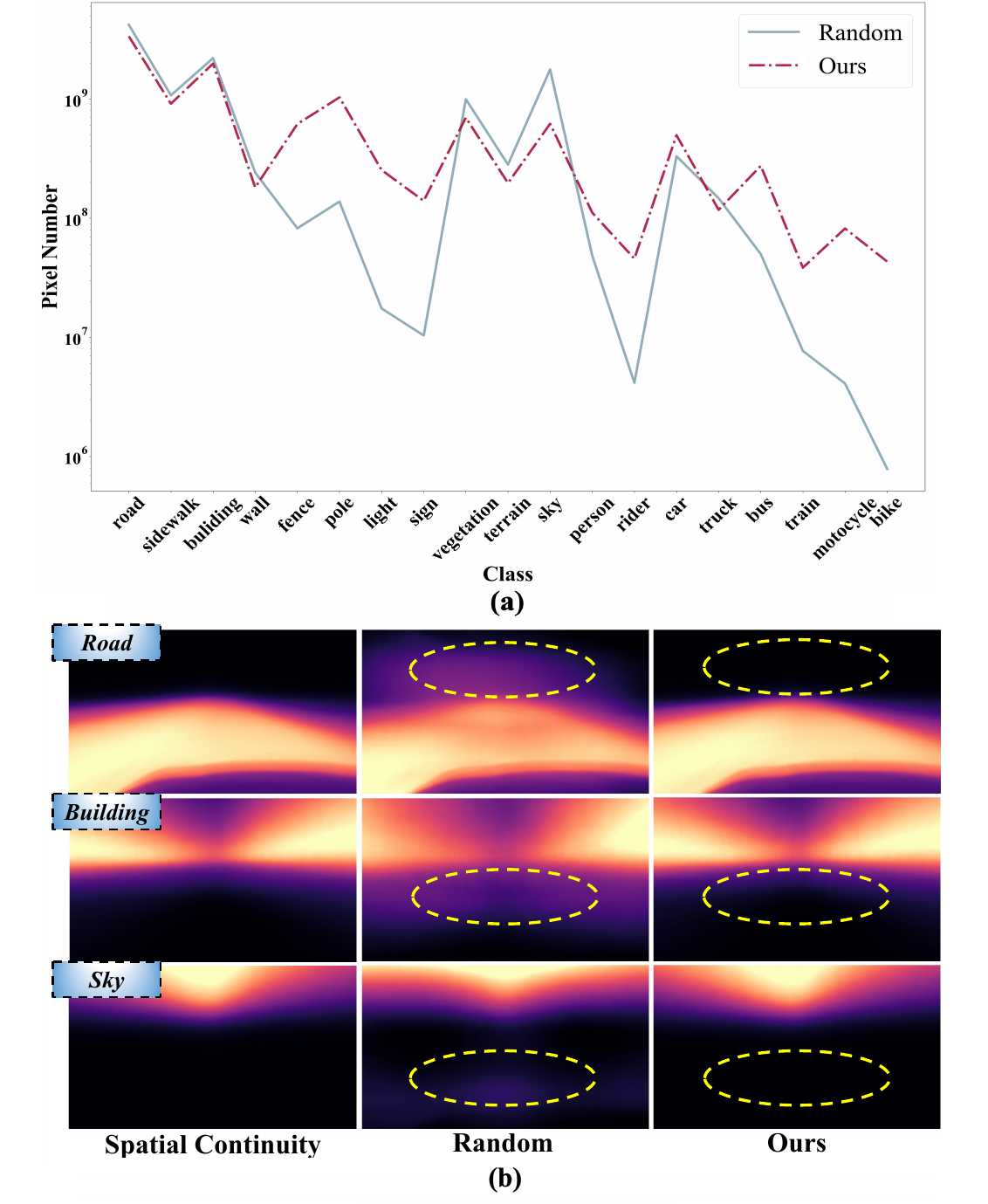}
  \caption{Visualization of class distribution and spatial continuity kernel in BDM. (a) shows comparison on the imbalance ratio of class-wise supervision. For each classes, we report the number of supervisory pixels that used in the training. Our method provide more class-balanced learning signal. (b) is the result of comparing the random cutmix and ours with the spatial continuity kernel of BDM.
}
  \label{fig:cb_sp_vis}
\end{figure}

\figref{fig:cb_sp_vis}-(a) shows the number of pixel-level supervision by class with and without patch level oversampling considering CB.
As shown in the figure, if patch level oversample is applied (\textit{e.g.} in GTA5, the probability that \textit{bike} will be selected is 23\%), it can be seen that the imbalance problem of pixel-level supervision is greatly alleviated.
~\figref{fig:cb_sp_vis}-(b) shows the spatial continuity of random cutmix and BDM. As shown in the figure, our proposed BDM can confirm that spatial continuity is maintained.

\paragraph{Limitations of Bidirectional Domain Mixup.}
Since BDM uses pseudo labels of the source and target dataset, the performance of the warm-up model greatly affects performance of the final model.
For example, as shown in ~\tableref{table:gta5}, IoU and mIoU differ greatly in the warm-up model.
However, it showed consistent performance improvement in representative UDA methods, such as entropy minimization~\cite{vu2019advent}, adversarial training~\cite{tsai2018learning}, and self-training~\cite{mei2020instance, zhang2021prototypical}.

\section{Conclusions}
In this paper, we proposed Bidirectional Domain Mixup (BDM), a cutmix method that cut the low confidence region and selects a patch to paste according to class-balance(CB), spatial continuity(SC), and pseudo label confidence(PC) in the corresponding region.
We observed that there are transferable and non-transferable regions in the source dataset, and showed that the non-transferable region can be easily removed through the pseudo label of the warm-up model.
Finally, we have shown through extensive experiments that CB, SC, and PC are important components for the paste mechanism in the context of UDA.
Our proposed BDM achieves state-of-the-art in GTA5 to cityscapes benchmark and SYNTHIA to cityscapes benchmark with a large gap.
We hope that our research will be applied not only to UDA but also to related fields, such as semi-supervised domain adaptation, and semi-supervised segmentation fields.

\section{Acknowledgements}
This research was supported by the the National Research Foundation of Korea(NRF) grant funded by the Korea government(MSIT) (No.2022R1F1A1075019, No. 2021M3E8A2100446).

\bigskip

\bibliography{aaai23}

\end{document}